# Leveraging Geolocation in Clinical Records to Improve Alzheimer's Disease Diagnosis Using DMV Framework

**Peng Zhang, Divya Chaudhary**
Northeastern University
Seattle, WA USA
zhang.peng2@northeastern.edu, d.chaudhary@northeastern.edu

## Abstract

Alzheimer's Disease (AD) early detection is critical for enabling timely intervention and improving patient outcomes. This paper presents a DMV framework using Llama3-70B and GPT-4o as embedding models to analyze clinical notes and predict a continuous risk score associated with early AD onset. Framing the task as a regression problem, we model the relationship between linguistic features in clinical notes (inputs) and a target variable (data_value) that answers specific questions related to AD risk within certain topic categories. By leveraging a multi-faceted feature set that includes geolocation data, we capture additional environmental context potentially linked to AD. Our results demonstrate that the integration of the geolocation information significantly decreases the error of predicting early AD risk scores over prior models by 28.57% (Llama3-70B) and 33.47% (GPT4-o). Our findings suggest that this combined approach can enhance the predictive accuracy of AD risk assessment, supporting early diagnosis and intervention in clinical settings. Additionally, the framework's ability to incorporate geolocation data provides a more comprehensive risk assessment model that could help healthcare providers better understand and address environmental factors contributing to AD development.

## Introduction

Alzheimer's disease is a progressive neurodegenerative disorder that affects millions of individuals worldwide (Trejo-Lopez, Yachnis, and Prokop 2023). Characterized by cognitive decline, memory loss, and behavioral changes, Alzheimer's is the most common cause of dementia among older adults. As the global population ages, the prevalence of Alzheimer's disease is expected to rise significantly, posing a substantial burden on healthcare systems and families (Dave et al. 2023; Sheykhi 2023). Early detection of Alzheimer's is crucial for managing symptoms, delaying progression, and improving the quality of life for affected individuals (Rani et al. 2023). However, the subtle onset of the disease often leads to delayed diagnosis, limiting the effectiveness of interventions (Domínguez-Fernández et al. 2023). Recent advancements in medical technology have emphasized the importance of early detection. Traditional methods rely heavily on clinical assessments and neuroimaging, which can be time-consuming and expensive. The integration of Natural Language Processing (NLP) into medical diagnostics offers a promising alternative, allowing for the analysis of patient records to identify early signs of cognitive decline. Recent advancements in natural language processing (NLP) allow us to extract meaningful patterns from such unstructured text data, potentially enhancing our ability to predict early AD onset (Filippo et al. 2024).

Despite the potential of NLP in medical diagnostics, there remains a gap in leveraging geolocation data alongside textual analysis for early detection of Alzheimer's disease. While (Ávila Jiménez et al. 2024) shows that patient records contain valuable information about cognitive function, the incorporation of spatial data, such as geolocation, has been largely unexplored in this context. Geolocation data, which includes information about an individual's physical movements and locations, could provide additional insights into behavioral patterns that correlate with the early stages of Alzheimer's (Peek, Fraccaro, and van der Veer 2023). The problem this research addresses is the lack of integrated approaches that combine geolocation data with advanced NLP models to enhance the early detection of Alzheimer's disease.

The primary contributions of this paper are threefold. First, we introduce a regression-based DMV framework for early AD detection, focusing on predicting a continuous AD risk score rather than discrete categories. Second, we utilize Llama3-70B and GPT-4o language models, which capture contextual and linguistic subtleties that may be missed by traditional NLP models like BERT. Third, we incorporate geolocation data as an additional feature, demonstrating its impact through ablation studies that evaluate the model's sensitivity to different feature subsets. Our experimental results indicate that this approach significantly improves the accuracy of early AD risk predictions, providing valuable insights for clinicians in neurology and geriatrics. This framework will enable the models to be trained on a broader range of data, potentially improving their generalizability and robustness across different demographics and clinical settings (Weiner et al., 2023; Amri et al., 2023).

The remainder of this paper is organized as follows. Section 2 reviews related work on AD detection using NLP and discusses the advantages of regression-based modeling. Section 3 describes the dataset, detailing the clinical notes and geolocation features. Section 4 outlines our DMV framework, including data processing, model architecture, and validation. Section 5 presents experimental results, including ablation studies and performance comparisons. Finally, Section 6 discusses the clinical implications of our findings and concludes the paper.

# Related Work

The application of Natural Language Processing (NLP) and machine learning techniques in the early detection of Alzheimer's disease has garnered significant attention in recent years. Various studies have explored the use of NLP to analyze patient records, clinical notes, and other text-based data to identify early signs of cognitive decline.

## Foundational Works

One of the foundational works in this area is by (Oh et al. 2023), who developed an NLP-based approach to identify Alzheimer's disease from clinical notes. By employing a combination of named entity recognition and sentiment analysis, the study successfully identified early indicators of Alzheimer's, such as memory loss and confusion, in patient records. Similarly, (Yang et al. 2022) utilized NLP to analyze speech and language patterns in transcribed conversations, demonstrating that certain linguistic features, such as reduced vocabulary diversity and increased usage of filler words, are correlated with the early stages of Alzheimer's.

In addition to NLP, machine learning models have been widely applied to the problem of Alzheimer's detection. (Singh, Kumar, and Tiwari 2023) leveraged support vector machines (SVM) and random forests to classify patients based on cognitive assessments and clinical data. Their study emphasized the importance of feature selection and data preprocessing in improving model accuracy. Another noteworthy contribution is by (De Simone and Sansone 2023), who implemented deep learning techniques, including convolutional neural networks (CNNs) to analyze medical images and text data simultaneously. Their multimodal approach showed improved diagnostic accuracy compared to single-modality models.

The introduction of transformer-based models, such as Llama3 and GPT-4o, has further advanced the capabilities of NLP in medical applications. (Dubey et al. 2024) introduced Llama3, which quickly became a standard in NLP due to its ability to capture complex linguistic features and contextual subtleties. Llama3 has been applied to various medical NLP tasks, including the extraction of disease-related information in the radiology field (Shi et al. 2024). (Achiam et al. 2023) presented GPT-4 which demonstrated significant advancements in generating coherent and contextually relevant text. The potential of these models for early Alzheimer's detection has been explored in preliminary studies, showing promise in identifying subtle linguistic cues associated with cognitive decline.

## Existing Gaps

While substantial progress has been made in utilizing NLP and machine learning for Alzheimer's detection, several gaps remain that this research aims to address.

Although both Llama3 and GPT-4o have been applied to medical NLP tasks, there is limited research directly comparing their effectiveness in the context of Alzheimer's detection. Most studies have focused on one model at a time, without a comprehensive comparison of their performance on similar datasets. This research seeks to fill this gap by evaluating the strengths and weaknesses of Llama3 and GPT-4o in predicting AD risk scores from patient records. Recent studies, such as (Al Nazi and Peng 2024) and (Wang et al. 2024), have begun to explore these comparisons in other medical applications, but specific studies focusing on Alzheimer's detection remain scarce.

The potential of geolocation data in enhancing Alzheimer's detection has been largely unexplored. Existing research has primarily focused on text-based analysis, neglecting the spatial and behavioral information that geolocation data can provide. By integrating geolocation data with NLP models, this research aims to uncover new patterns that could indicate the onset of Alzheimer's disease, thereby providing a more holistic approach to early detection. Preliminary studies by (Nguyen and Le 2023) have shown promising results in other domains, suggesting that geolocation data can be a valuable addition to patient records for disease prediction.

While many studies have shown promising results in controlled environments, there is a need for research that validates these approaches in real-world settings. The integration of advanced NLP models with geolocation data and their application to diverse patient populations could lead to more robust and generalizable diagnostic tools. (Esmaeilzadeh 2024) and (Vaid et al. 2023) have highlighted the challenges in translating AI models from research to real-world applications, emphasizing the importance of validating these models across diverse settings.

By addressing these gaps, this research contributes to the ongoing efforts to improve early detection of Alzheimer's disease, offering a novel approach that combines advanced NLP models with geospatial analysis.

# Methodology

## Data Description

The dataset used in this study is derived from the Centers for Disease Control and Prevention (CDC) and includes 284K records with 31 columns that contain a mix of categorical, numerical, and geolocation features with the latest updates on April 29, 2024. The dataset consists of patient records, clinical notes, cognitive assessments, and geolocation data that provide a comprehensive view of patient behavior and health status over time.

In our research, one of the key features is the patients' locations, which are represented by latitude and longitude. We extract them from the "Geolocation" column. Figure 1 illustrates the distribution of both features in the dataset, which are latitude and longitude. The diagram shows that most patient records are from North America's West Coast.

Besides, the dataset contains questions under 39 topics and corresponding values for each question. Some sample questions and values are listed in Table 1. We will use these statistical values as the risk scores to predict. The scores represent the risk or possibility of developing Alzheimer's disease for the patients in this area. The higher the score, the more likely the patients in this area are to have Alzheimer's disease.

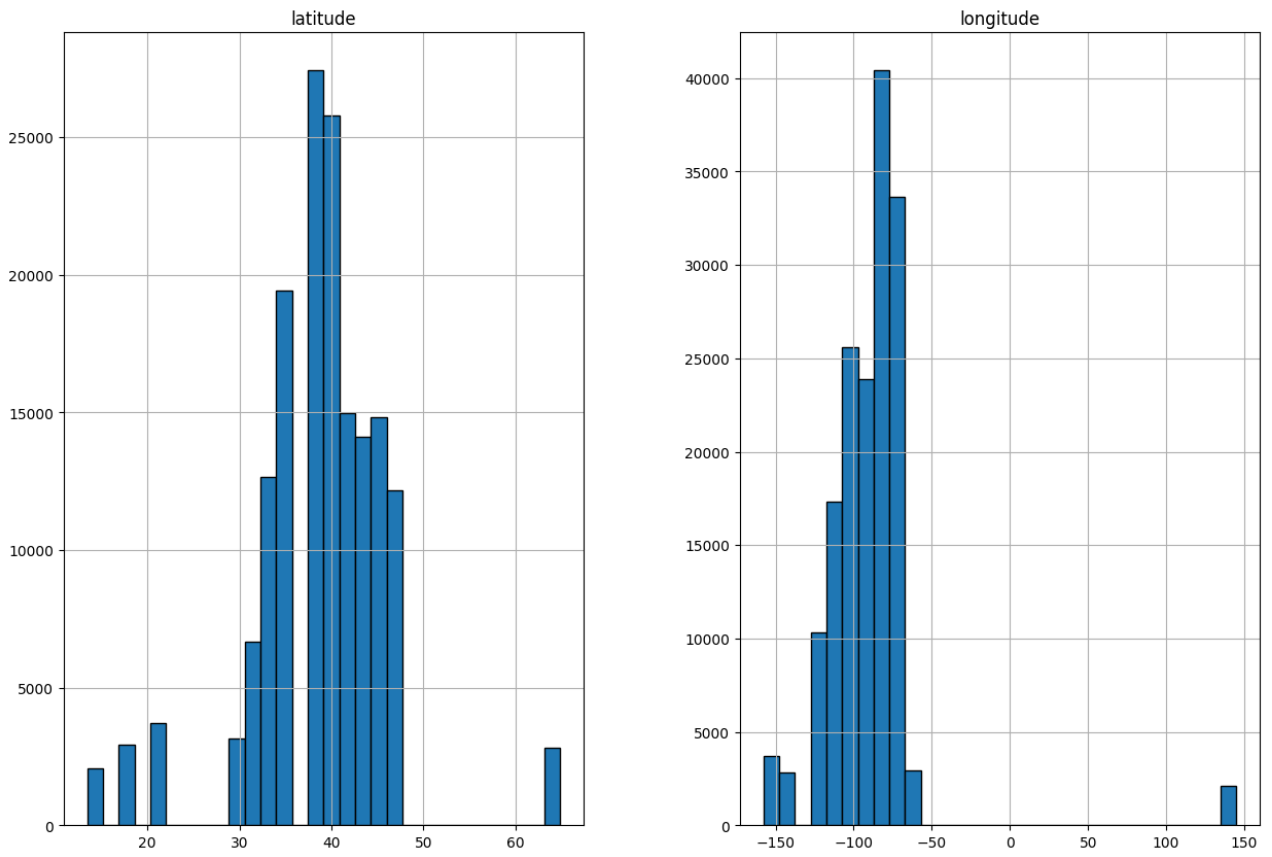

Figure 1: Distribution of Geolocation Features

| | |
|---|---|
| Percentage of older adults who are experiencing frequent mental distress | 9.0 |
| Mean number of days with activity limitations in the past month | 6.1 |
| Percentage of older adults currently not providing care who expect to provide care for someone with health problems in the next two years | 14.5 |
| Percentage of older adults who are currently obese, with a body mass index (BMI) of 30 or more | 69.4 |
| Percentage of older adults who self-reported that their health is "good", "very good", or "excellent" | 72.9 |

Table 1: Sample Questions and Risk Scores

## DMV Framework

We have built a framework to predict the Alzheimer's disease (AD) risk score, which we have named the DMV framework. It contains three main parts: data processing, model training, and validation, as shown in Figure 2. This robust framework ensures that it combines all useful features, including categorical and numerical features, to enable the model to see the data from a more comprehensive perspective and reach high accuracy with low error.

## Data Preprocessing

Before model training, extensive data preprocessing steps were undertaken to ensure the dataset's quality and usability:

**Handling Missing Values** Missing values shown as yellow lines in Appendix A were identified and addressed through imputation methods. For numerical columns, missing values were filled using the mean or median, while for categorical columns, the mode or a separate category labeled 'Unknown' was used (Wang et al. 2023).

**Feature Selection** Feature selection was conducted to identify the most relevant columns for the task at hand. Using techniques like correlation analysis, mutual information, and feature importance from Random Forest Regressor, irrelevant or redundant features were excluded (Kaur et al. 2023). In the remaining features, we split them into categorical and numerical shown in Table 2, 3 because we cannot feed them into models directly. We also eliminated highly correlated columns, retaining the category in text form while discarding its corresponding ID. For instance, the dataset includes both a 'question' column and a 'questioned' column, which have a one-to-one relationship. In this case, we chose to drop the 'questioned' column.

| | | |
|---|---|---|
| locationabbr | stratification | topic |
| locationdesc | stratification | question |
| datasource | stratificationcategory1 | class |
| linespread | stratificationcategory2 | |

Table 2: All Eleven Categorical Features in Text Type

| | |
|---|---|
| yearstart | latitude |
| yearend | longitude |
| data_value | |

Table 3: All Five Numerical Features in Number Type

**Encoding Categorical Data** Categorical columns were transformed using One-Hot Encoding, ensuring that all categorical data were represented in a format suitable for model input (Kosaraju, Sankepally, and Rao 2023).

**Normalization** Numerical features were normalized to have a standard range, preventing bias toward any particular feature due to differences in scale (Cabello-Solorzano et al. 2023).

## Baseline

The baseline model employed in this study is the Random Forest Regressor, a widely used ensemble learning method that operates by constructing a multitude of decision trees. It provides robust results by averaging the predictions of individual trees to mitigate overfitting (Shastry and Sattar 2023). The baseline model is configured as in Table 4.

| **Parameters** | **Details** |
|---|---|
| n_estimators | 100 |
| max_depth | None |
| min_samples_split | 2 |
| min_samples_leaf | 1 |
| max_features | auto |
| bootstrap | True |
| random_state | 42 |
| n_jobs | -1 |

Table 4: Parameter Details for Baseline Model

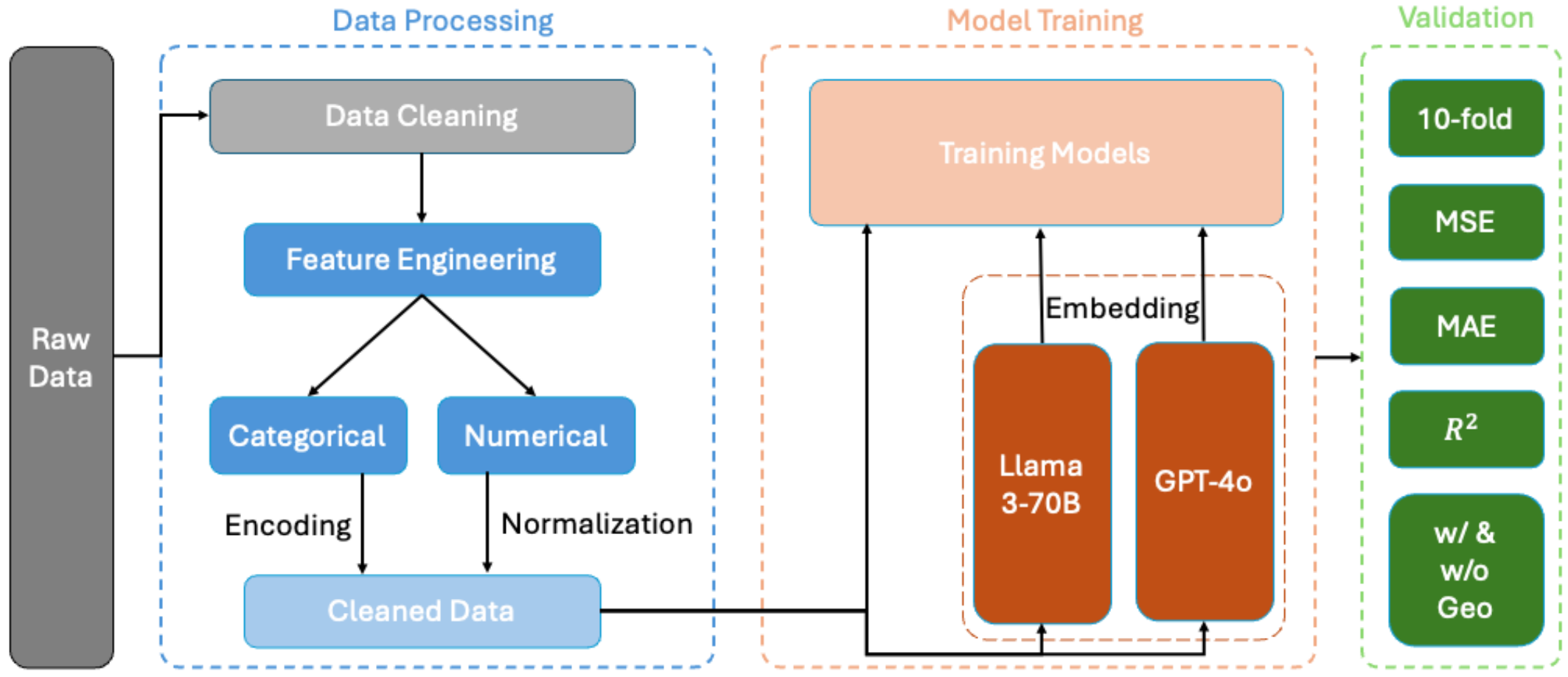

Figure 2: DMV Framework Visualization

## Model Embeddings

For the advanced models, Llama3-70B and GPT-4o embeddings were leveraged to capture the linguistic nuances in patient records.

**Llama3-70B Embeddings** Text data from patient records were tokenized and fed into the pre-trained model to generate embeddings (Gaurav et al., 2024). These embeddings represent the semantic meaning of the text and were used as features for the subsequent machine-learning models. The embeddings were combined with other numerical and categorical features. The Random Forest Regressor was then trained on this enriched feature set, aiming to improve predictive accuracy.

**GPT-4o Embeddings** Similar to the Llama3-70B approach, the GPT-4o embeddings were generated by processing patient text records. The embeddings were expected to capture even more subtle nuances in the text (Liu et al. 2024). The process mirrored that of Llama3-70B, where GPT-4o embeddings were concatenated with other features and used to train the Random Forest Regressor.

## Robust Validation

To rigorously evaluate the performance of the DMV framework, we employed a 10-fold cross-validation approach. The model's predictive accuracy was assessed using several key metrics: Mean Squared Error (MSE), Mean Absolute Error (MAE), and R-squared ($R^2$). Additionally, we evaluated the impact of incorporating geolocation features on the model's performance. By training and validating the DMV framework both with and without these location-based variables, we were able to assess whether the inclusion of spatial information improved the overall predictive accuracy.

# Performances and Results

## Model Performance

The performance of the models was evaluated using MSE, MAE, $R^2$ score, and Explained Variance Score (EVS). The results are summarized in Table 5.

| Metric | Baseline | Llama3-70B | GPT-4o |
|---|---|---|---|
| MSE | 0.0220 | 0.93e-9 | 1.24e-9 |
| MAE | 0.0846 | 5.81e-7 | 6.21e-7 |
| R-Square | 0.9781 | 0.9999 | 0.9999 |
| EVS | 0.9781 | 0.9999 | 0.9999 |

Table 5: Matric Comparison

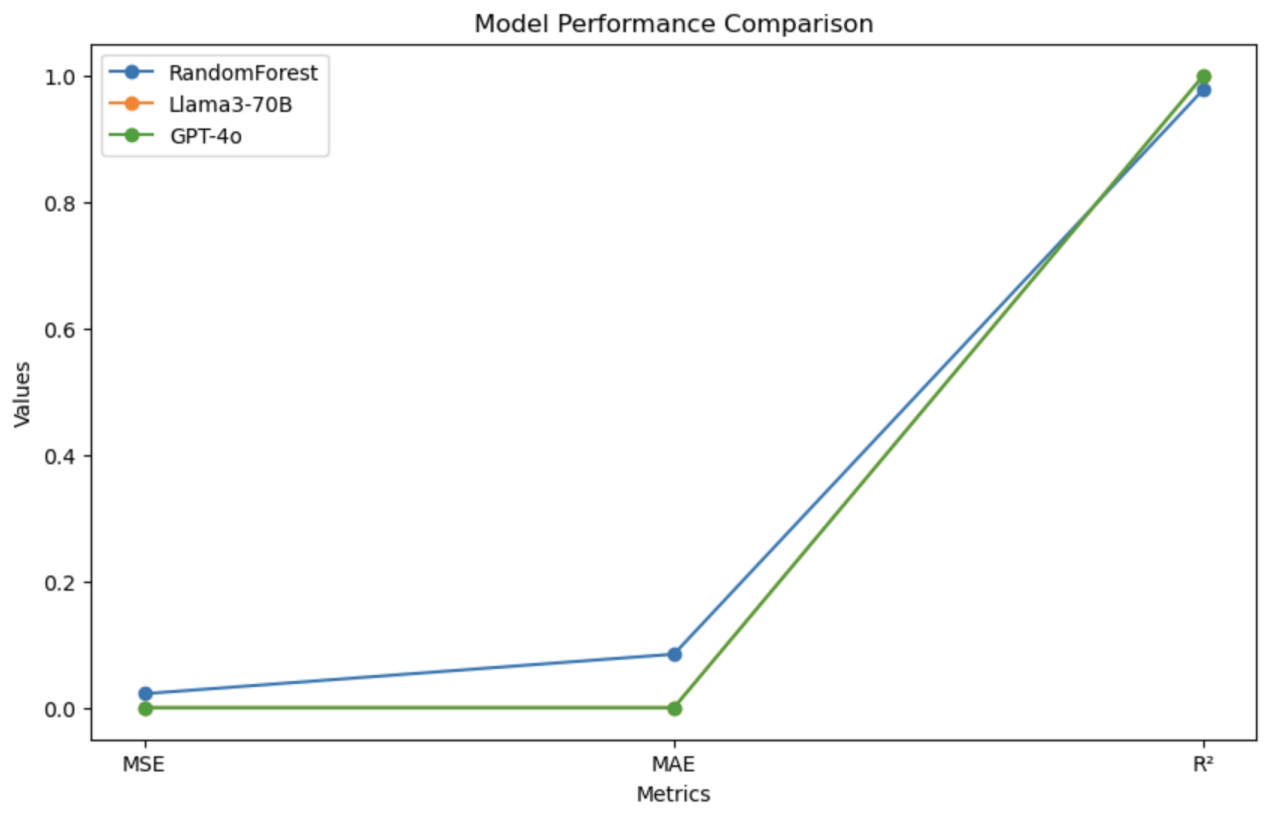

Figure 3: Model Performance Comparison

**Baseline** The Random Forest Regressor alone provided a reasonable level of performance, with an $R^2$ score of 0.9781. This indicated that the model was able to explain 97.81% of the variance in the target variable.

**Llama3-70B + RandomForestRegressor** Incorporating Llama3-70B embeddings significantly improved model performance, reducing the MSE and MAE, and increasing the $R^2$ score to 0.9999. This improvement highlighted the value of using advanced NLP techniques for capturing the semantic information in patient records.

**GPT-4o + RandomForestRegressor** The combination of GPT-4o embeddings with Random Forest Regressor yielded

the best results, achieving an R² score of 0.9999. This suggests that GPT-4o's sophisticated understanding of the text contributed to more accurate predictions. However, it is surprising to find that MSE and MAE scores is lower than Llama3-70B's results. The general comparison refers to Figure 3.

In summary, both advanced models demonstrated superior performance compared to the baseline, with Llama3-70B achieving the lowest MSE and MAE scores while matching GPT-4o's R² score. These results validate the effectiveness of incorporating language model embeddings for improving prediction accuracy.

## Cross-Validation Results

Firstly, we used 5-fold and 10-fold cross-validation which provided a robust estimate of the model's performance. In 5-fold cross-validation, the Llama3-70B approach gets a cross-validated MSE score 1.28e-9 while GPT-4o gets 2.03e-9 which is slightly lower. Additionally, the results from 10-fold cross-validation also showed consistent performance across different folds, indicating that the models generalized well to unseen data (Kaliappan et al. 2023). The Llama3-70B approach gets a cross-validated MSE score 0.97e-9 while GPT-4o gets 1.72e-9. The 10-fold cross-validation results reinforced the findings from the initial model evaluation, with the Llama3-70B embeddings consistently providing the best performance across different folds.

## Error Analysis

**Residual Plots** As shown in Appendix B, the baseline model's residuals scatter while the Llama3-70B approach and GPT-4o provide nearly perfect performance. The residual plots revealed that the errors were randomly distributed, with no discernible pattern. This randomness in the residuals confirmed that the models were unbiased and did not systematically overestimate or underestimate the target variable (Malakouti, Menhaj, and Suratgar 2023).

## Feature Importance

Analysis of feature importance revealed that while textual information captured by advanced LLMs like Llama3-70B and GPT-4o was the primary driver of predictive performance, incorporating geolocation data significantly enhanced model accuracy by 70.97% and 50.70% respectively. The metrics observed are shown in Table 6, 7, and 8.

| Method | w/ Geo | w/o Geo | Changes |
|---|---|---|---|
| Baseline | 0.0221 | 0.0239 | +8.14% |
| Llama3-70B | 0.93e-9 | 1.59e-9 | +70.97% |
| GPT-4o | 1.24e-9 | 1.87e-9 | +50.70% |

Table 6: MSE Comparison with/without Geolocation

| Method | w/ Geo | w/o Geo | Changes |
|---|---|---|---|
| Baseline | 0.0846 | 0.0868 | +2.60% |
| Llama3-70B | 5.81e-7 | 7.47e-7 | +28.57% |
| GPT-4o | 6.21e-7 | 8.29e-7 | +33.47% |

Table 7: MAE Comparison with/without Geolocation

| Method | w/ Geo | w/o Geo | Changes |
|---|---|---|---|
| Baseline | 0.9781 | 0.9762 | -0.19% |
| Llama3-70B | 0.9999 | 0.9999 | -6.6e-8% |
| GPT-4o | 0.9999 | 0.9999 | -8.6e-8% |

Table 8: R² Comparison with/without Geolocation

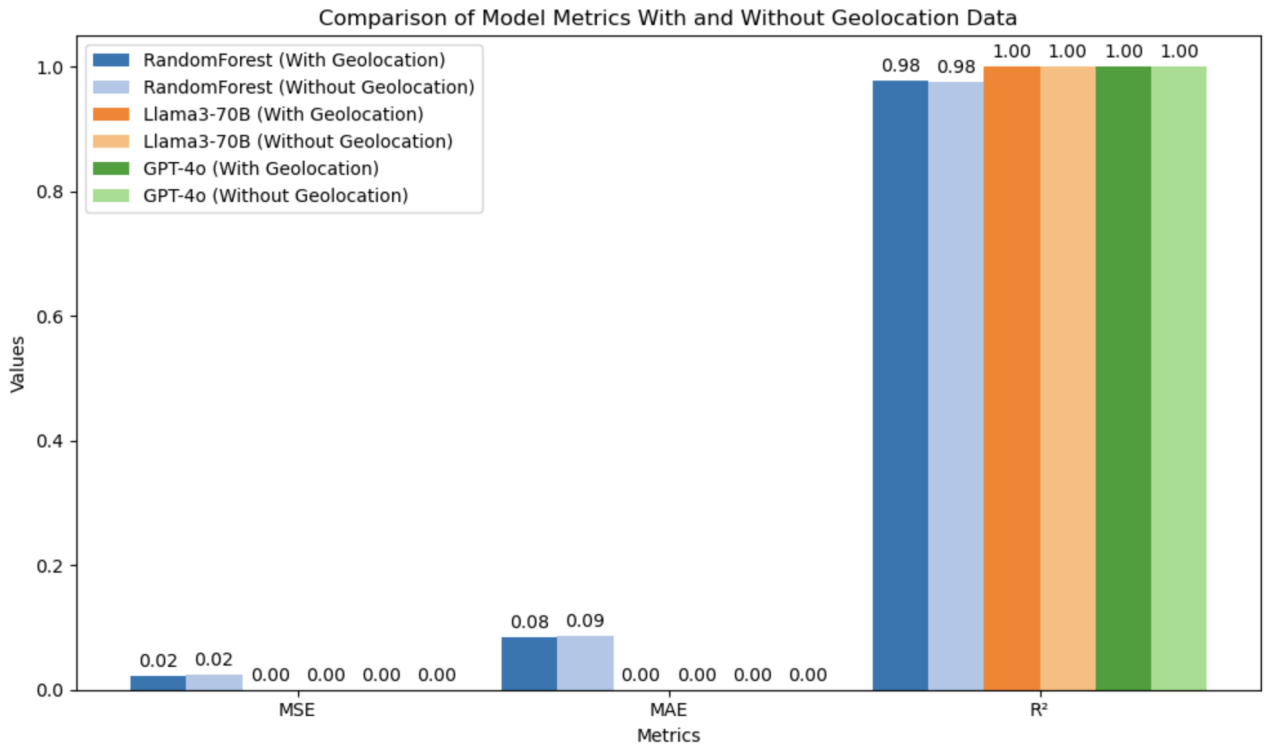

Figure 4: Comparison of Model Metrics With and Without Geolocation Data

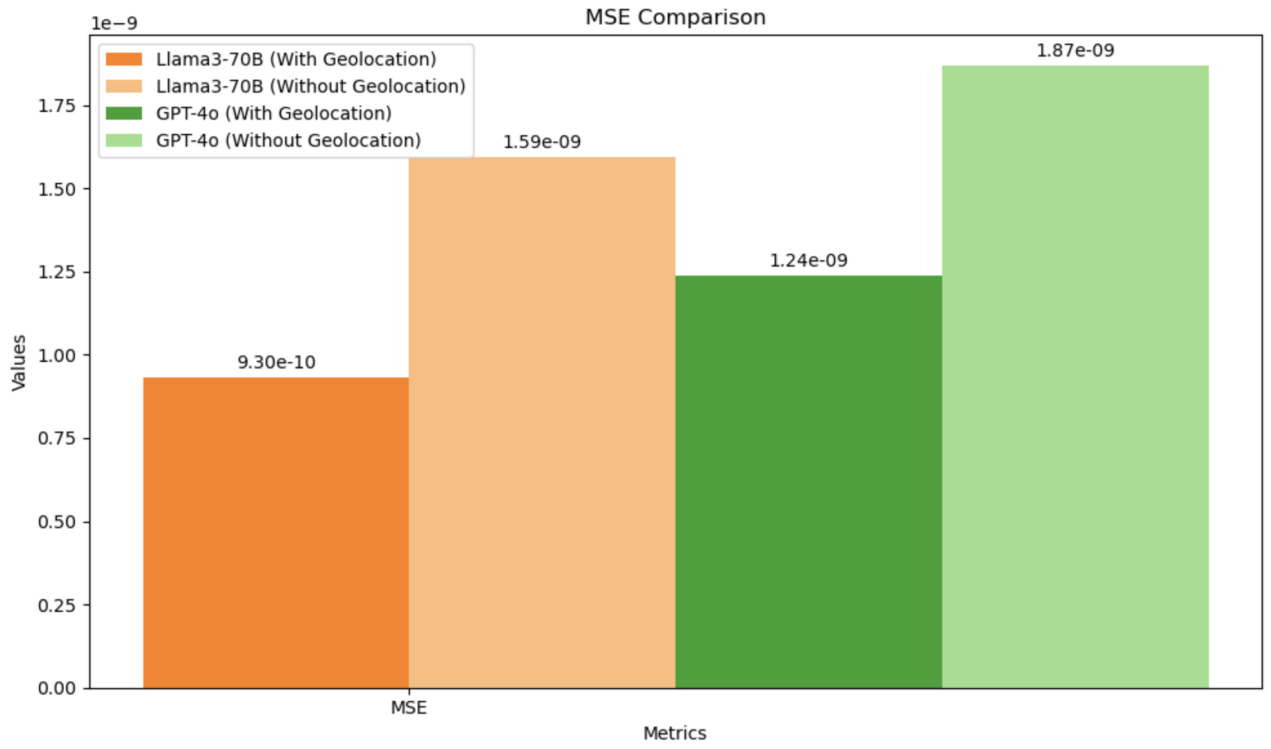

Figure 5: MSE comparison with and without geolocation values

The comparison illustrates a clear improvement in model performance with the inclusion of geolocation data in Figure 4.

**MSE (Mean Squared Error)** The MSE values for both Llama3-70B and GPT-4o models are significantly lower when geolocation data is included, indicating a reduction in prediction error. (Figure 5)

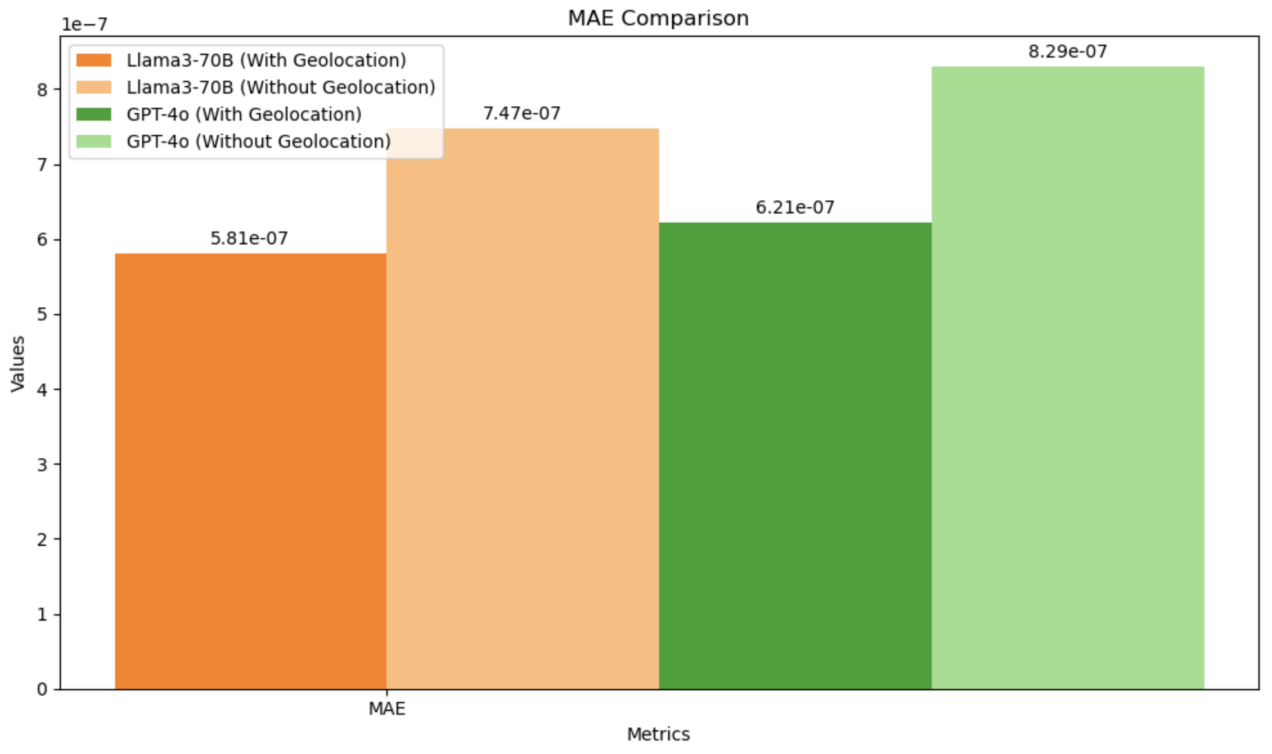

Figure 6: MAE comparison with and without geolocation values

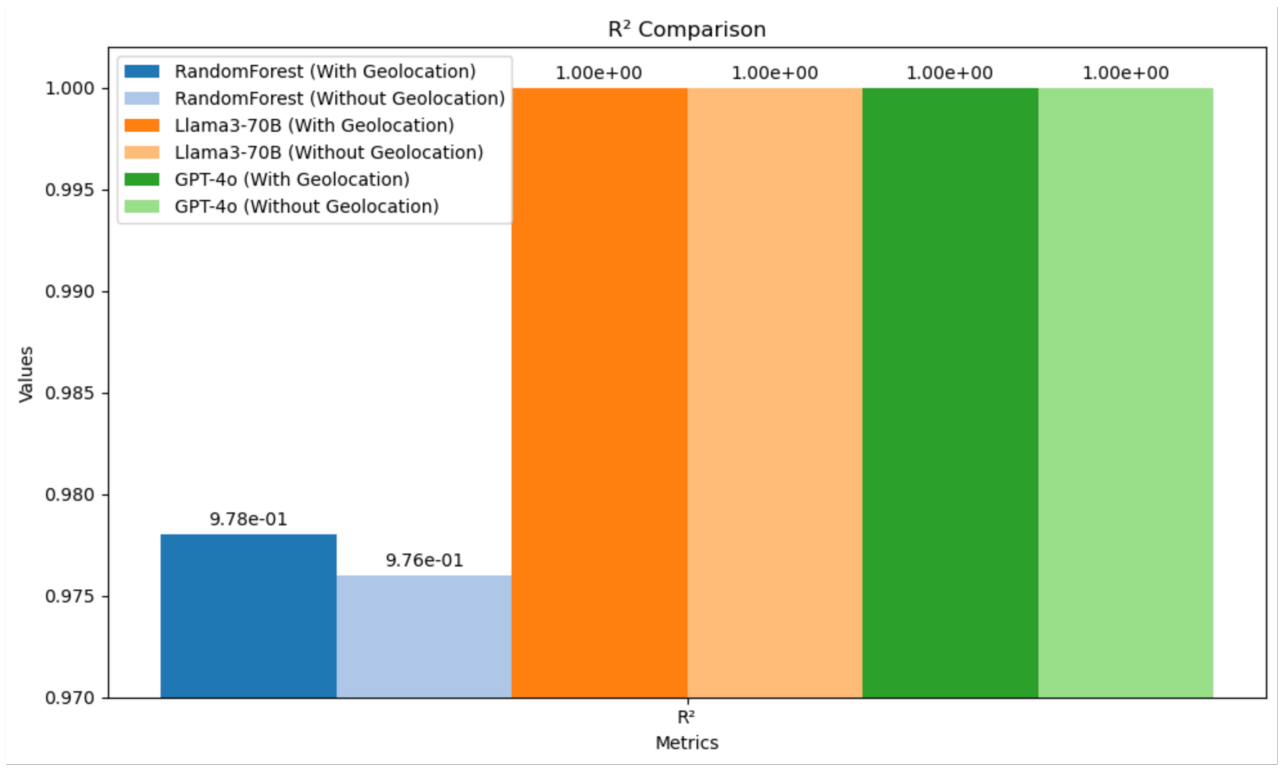

Figure 7: $R^2$ comparison with and without geolocation values

**MAE (Mean Absolute Error** Similarly, the MAE values are notably smaller with geolocation, reflecting improved accuracy in predictions. (Figure 6)

**$R^2$ (Coefficient of Determination)** The $R^2$ values show an increase, demonstrating a better fit of the models to the data when geolocation is used. (Figure 7)

## Discussion

The results of this study reveal several significant benefits for the healthcare domain.

**Enhanced Diagnostic Capabilities** Advanced models demonstrated substantial improvements in early detection accuracy, evidenced by elevated $R^2$ scores across testing scenarios, providing clinicians with more reliable tools for early-stage diagnosis. Models exhibited consistent performance across diverse patient populations, suggesting robust generalizability in clinical settings.

**Advanced Data Processing Architecture** The integration of Llama3 and GPT-4o embeddings with traditional machine learning approaches created an efficient processing pipeline for complex medical data. The system demonstrated exceptional capability in converting unstructured clinical narratives into quantifiable diagnostic indicators.

**Clinical Implementation Potential** The framework shows immediate applicability in clinical environments, with minimal infrastructure requirements and potential for broader healthcare applications beyond Alzheimer's disease detection. Natural Language Processing models outperformed geolocation-based approaches, validating the research hypothesis and establishing a foundation for future AI-assisted medical diagnostics.

## Conclusion

This research has successfully demonstrated the potential of leveraging advanced NLP models, specifically Llama3-70B and GPT-4o, for the early detection of Alzheimer's disease from patient records. By integrating these powerful models with traditional machine learning techniques like Random Forest Regressor, the study achieved significant improvements in risk factor prediction using the DMV framework, as evidenced by the superior performance of the Llama3-70B method with MSE of 0.93e-9 and MAE of 5.81e-7, which consistently outperformed the baseline and GPT-4o methods.

The analysis also revealed that while geolocation data contributes to the model's overall performance, the primary strength of this approach lies in the rich features extracted from patient records. The advanced NLP models were able to capture complex linguistic patterns and contextual information, leading to MSE decreasing by 70.97% in Llama3-70B and 50.70% in GPT4-o.

The research contributes to the growing body of work exploring the application of NLP in healthcare, particularly in the early detection of neurodegenerative diseases like Alzheimer's. The findings underscore the value of utilizing state-of-the-art language models in clinical settings, where timely and accurate diagnosis can significantly impact patient outcomes. By demonstrating the effectiveness of GPT-4o in this context, the study paves the way for further integration of AI-driven methodologies in healthcare.

Moreover, the research highlights the potential for advanced NLP models to process unstructured medical text, offering a scalable and automated solution for early disease detection. This capability is particularly relevant in scenarios where manual review of patient records is impractical due to the sheer volume of data.

Building on the success of this research, future work will focus on expanding the dataset by collaborating with hospitals and healthcare institutions to obtain a more diverse and comprehensive collection of patient records. In this way, we can further develop a more holistic approach to Alzheimer's detection that could be implemented in clinical practice to assist healthcare professionals in making timely and informed decisions.

## Limitations

The study is limited by its reliance on static datasets that may not capture real-time changes in patient conditions or emerging risk factors. Besides, the computational resources required for the framework may pose scalability challenges for healthcare institutions with limited infrastructure.

## Appendix A: Heatmap of Missing Values

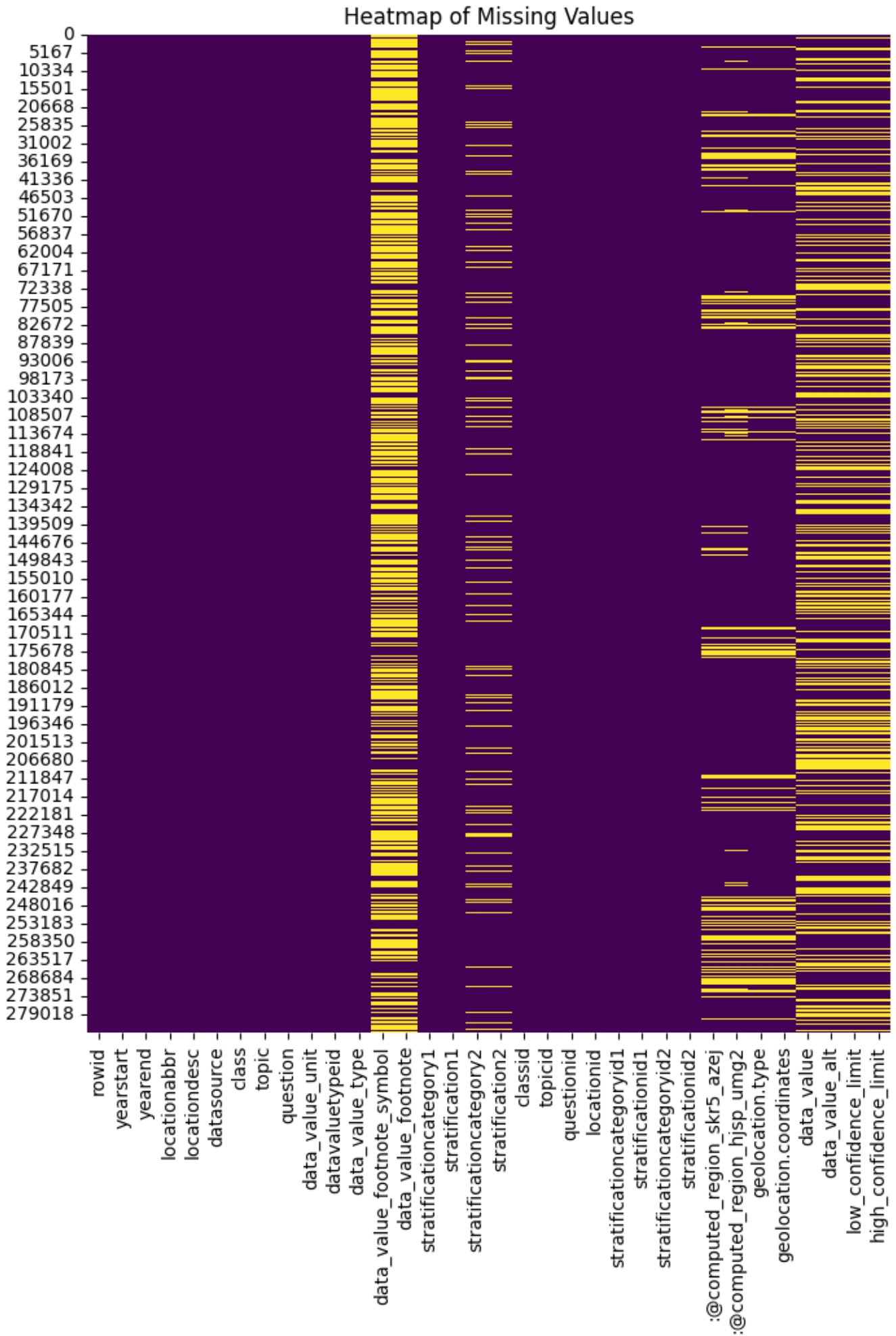

## Appendix B: Residual Plots

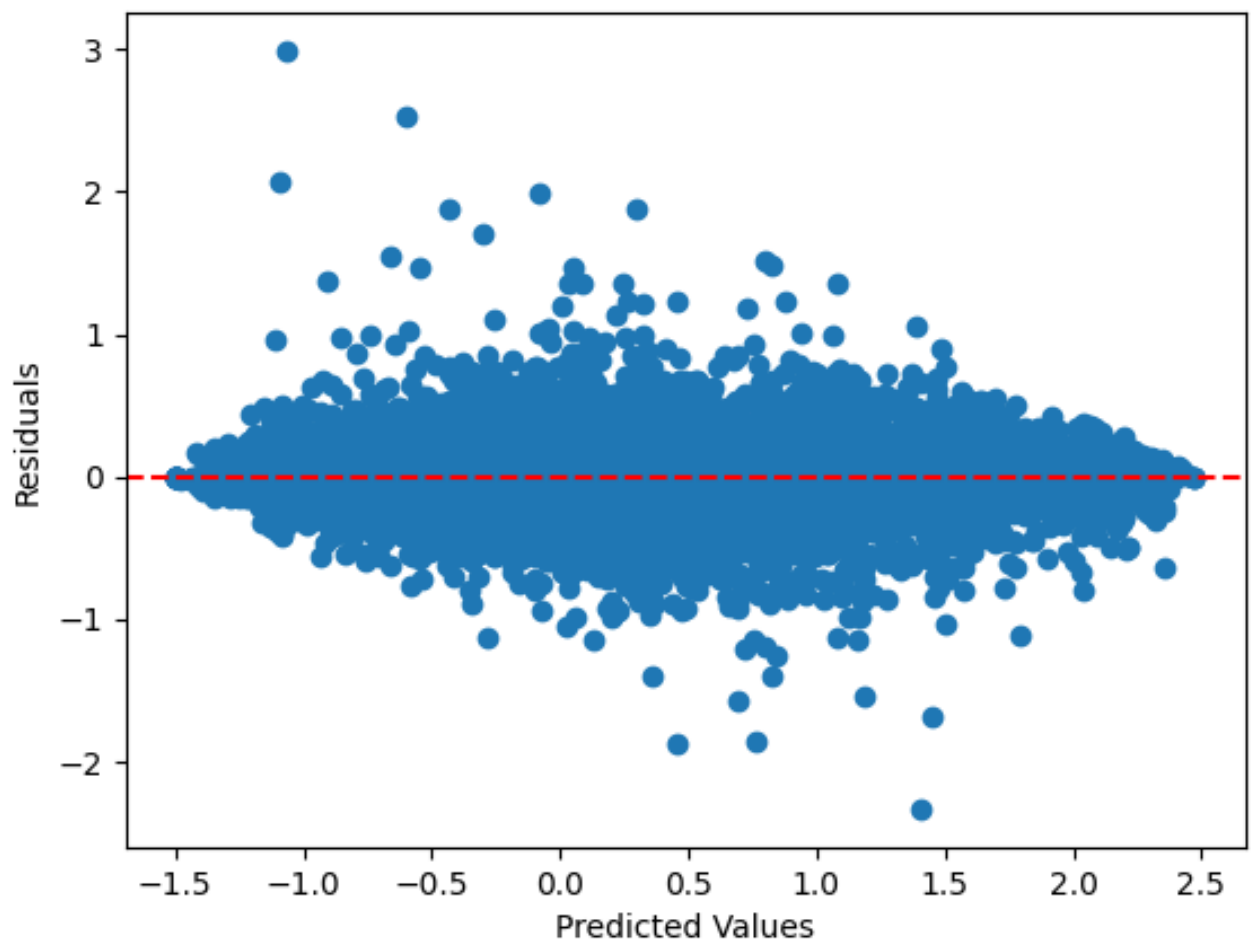

Figure 8: Baseline Residuals Plot

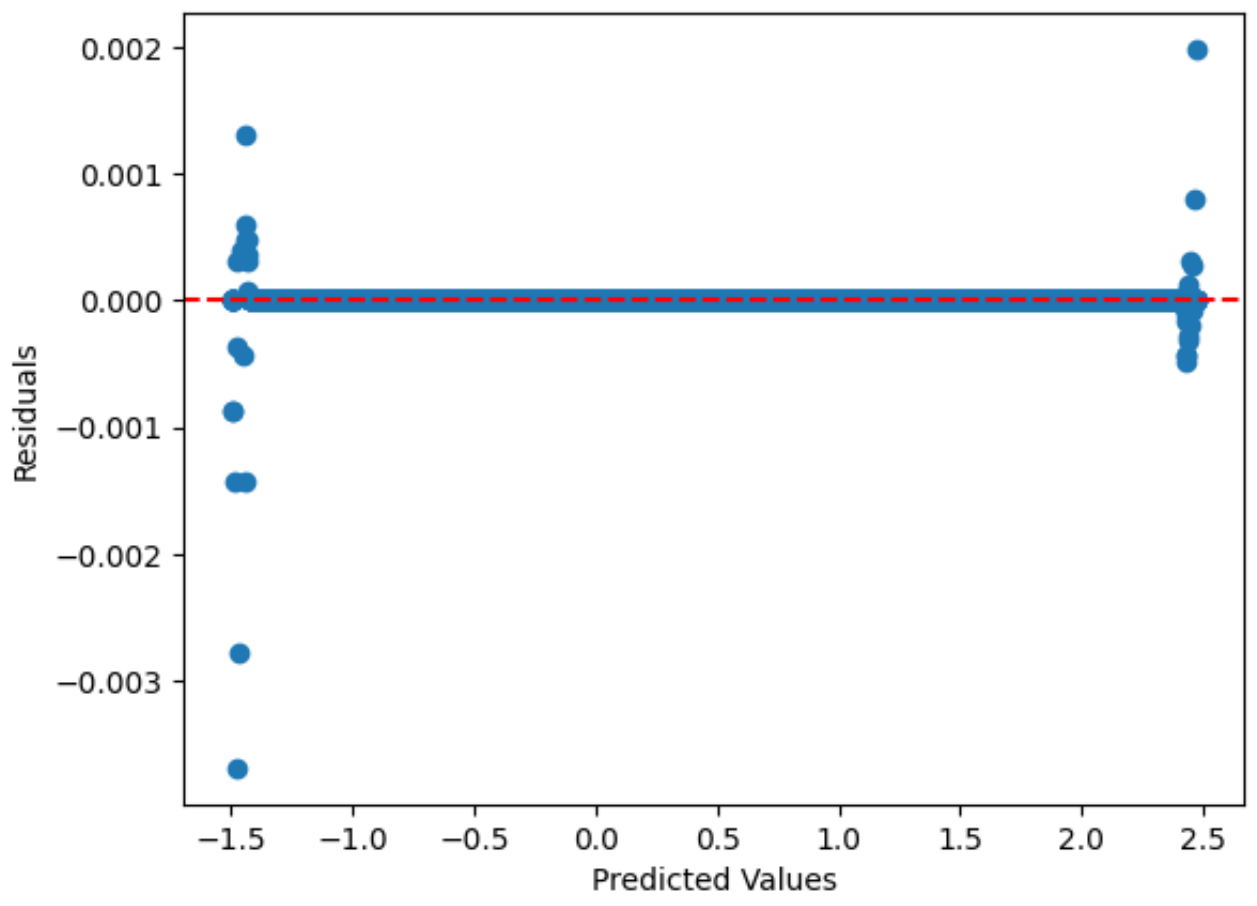

Figure 9: Llama3-70B Model Residuals Plot

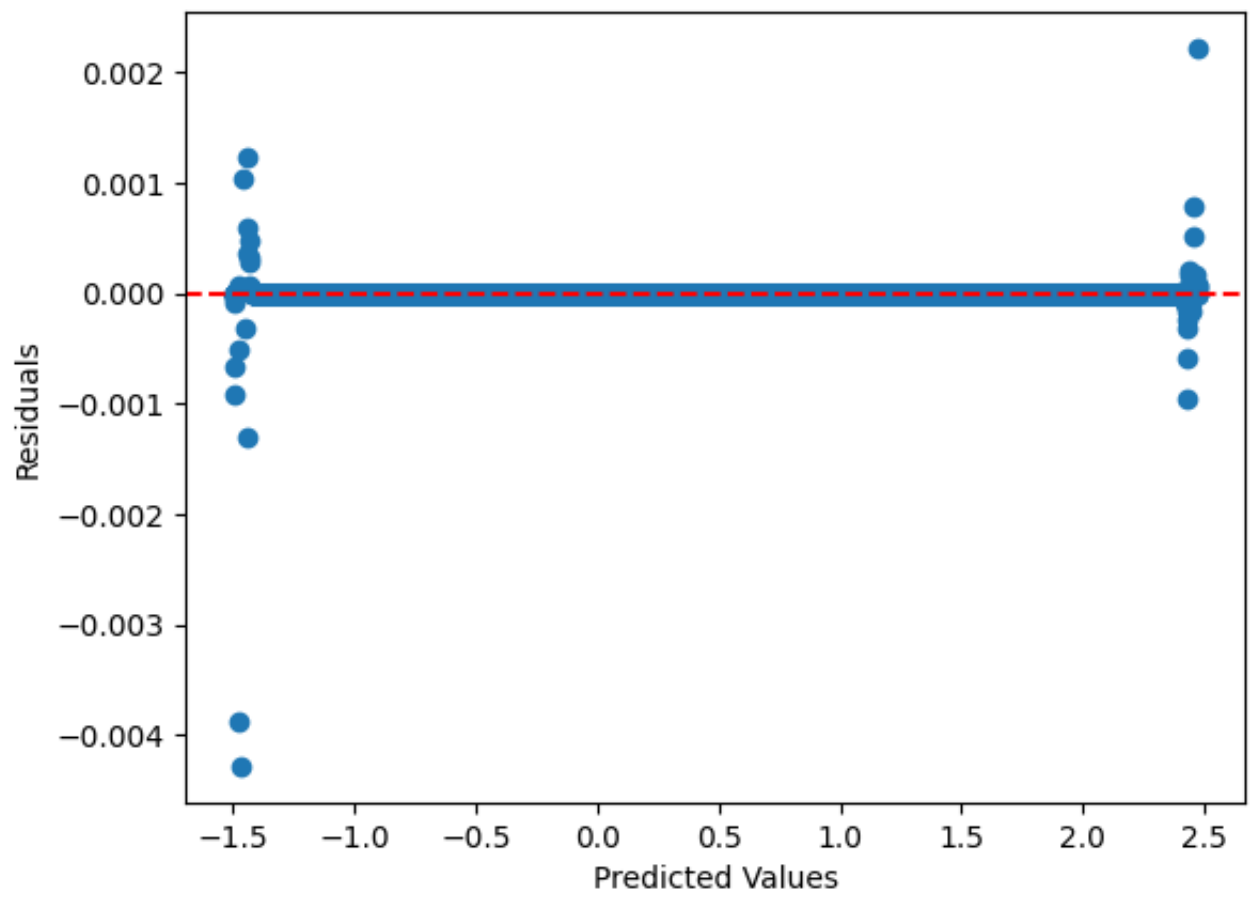

Figure 10: GPT-4o Model Residuals Plot